\documentclass[journal]{IEEEtai}

\usepackage[colorlinks,urlcolor=blue,linkcolor=blue,citecolor=blue]{hyperref}

\usepackage{color,array}

\usepackage{graphicx}
\usepackage{algorithm}
\usepackage{algorithmic}
\usepackage[switch]{lineno}
\usepackage{amsthm,amsmath,amssymb}
\usepackage{mathrsfs}
\usepackage{multirow}
\usepackage{makecell} 

\usepackage{booktabs}       
\usepackage{nicefrac}       
\usepackage{microtype}      
\usepackage{xcolor}         
\usepackage{subfig}         

\begin{document}

\title{Temporal Knowledge Sharing enable \\Spiking Neural Network \\Learning from Past and Future}

\author{Yiting Dong, Dongcheng Zhao, and Yi Zeng
\thanks{This work was supported in part by the Key Research Program of Frontier Sciences, CAS (Grant No. ZDBS-LY-JSC013). }
\thanks{Yiting Dong and Dongcheng Zhao contributed equally to this work.}

\thanks{Yiting Dong is with the School of Future Technology, University of Chinese Academy of Sciences, Beijing 100049, China, and also with the Brain-Inspired Cognitive Intelligence Lab, Institute of Automation, Chinese Academy of Sciences, Beijing 100190, China (e-mail:dongyiting2020@ia.ac.cn).}
\thanks{Dongcheng Zhao is with the Brain-Inspired Cognitive Intelligence Lab, Institute of Automation, Chinese Academy of Sciences, Beijing 100190, China (e-mail:zhaodongcheng2016@ia.ac.cn).}
\thanks{Yi Zeng is with the Brain-Inspired Cognitive Intelligence Laboratory, Institute of Automation, Chinese Academy of Sciences, Beijing 100190, China, with the School of Artificial Intelligence, University of Chinese Academy of Sciences, Beijing 100049, China, with the Center for Excellence in Brain Science and Intelligence Technology, Chinese Academy of Sciences, Shanghai 200031, China, also with State Key Laboratory of Multimodal Artificial Intelligence Systems, Institute of Automation, Chinese Academy of Sciences, Beijing, China 100190 (e-mail: yi.zeng@ia.ac.cn).}
\thanks{This paragraph will include the Associate Editor who handled your paper.}}

\markboth{Journal of IEEE Transactions on Artificial Intelligence, Vol. 00, No. 0, Month 2023}
{First A. Author \MakeLowercase{\textit{et al.}}: Bare Demo of IEEEtai.cls for IEEE Journals of IEEE Transactions on Artificial Intelligence}

\maketitle

\begin{abstract}

Spiking Neural Networks (SNNs) have attracted significant attention from researchers across various domains due to their brain-like information processing mechanism. However, SNNs typically grapple with challenges such as extended time steps, low temporal information utilization, and the requirement for consistent time step between testing and training. These challenges render SNNs with high latency. Moreover, the constraint on time steps necessitates the retraining of the model for new deployments, reducing adaptability.
To address these issues, this paper proposes a novel perspective, viewing the SNN as a temporal aggregation model. We introduce the Temporal Knowledge Sharing (TKS) method, facilitating information interact between different time points. TKS can be perceived as a form of temporal self-distillation. To validate the efficacy of TKS in information processing, we tested it on static datasets like CIFAR10, CIFAR100, ImageNet-1k, and neuromorphic datasets such as DVS-CIFAR10 and NCALTECH101. Experimental results demonstrate that our method achieves state-of-the-art performance compared to other algorithms.
Furthermore, TKS addresses the temporal consistency challenge, endowing the model with superior temporal generalization capabilities. This allows the network to train with longer time steps and maintain high performance during testing with shorter time steps. Such an approach considerably accelerates the deployment of SNNs on edge devices. Finally, we conducted ablation experiments and tested TKS on fine-grained tasks, with results showcasing TKS's enhanced capability to process information efficiently.
\end{abstract}

\begin{IEEEImpStatement}
Spiking Neural Networks have garnered significant interest across various domains due to their brain-like information processing mechanisms. The presence of spiking neurons endows SNNs with efficient spatio-temporal dynamics, making them particularly adept at handling data with temporal features. However, typically, SNNs face issues such as latency due to long time steps, low utilization of temporal information, and the need for consistency in steps between testing and training. These challenges have hindered the deployment of SNNs across devices.
With the introduction of our method, we have not only enhanced the ability of SNNs to process temporal features but also reduced the steps required during the testing phase. Our approach addresses the necessity for consistent steps between testing and training. It allows the SNN to achieve superior performance in shorter step compared to the training phase, significantly boosting the adaptability and ease of deployment of SNNs.
\end{IEEEImpStatement}


\section{Introduction}

\IEEEPARstart{S}{piking} Neural Networks (SNNs) have attracted widespread attention due to their excellent dynamical properties and their capability to process spatio-temporal information\cite{roy2019towards,kim2022neural,kim2021optimizing}. Compared to conventional artificial neural networks, spiking neurons in SNNs introduce a temporal-dependency capability by accumulating membrane potentials. They convey information by transmitting discrete spikes across temporal and spatial dimensions\cite{kim2022neural,kim2021optimizing}. The sequence of spikes transmitted within an SNN embodies the temporal correlations of information. SNNs extract relevant information by capturing the temporal frequency and the timing of the spikes.

\begin{figure}[h]
	\centering
	\includegraphics[width=1.0\columnwidth]{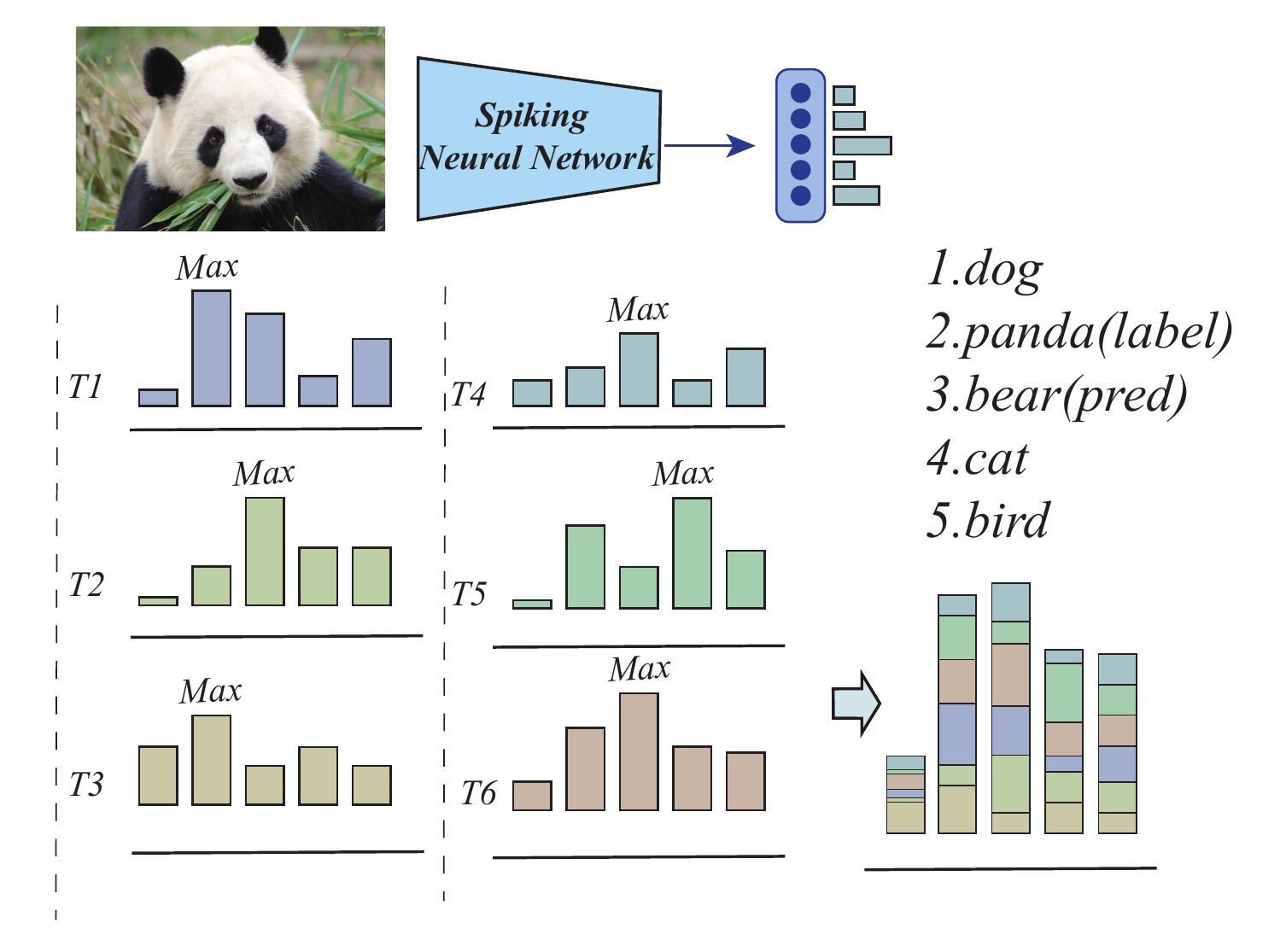} 
	\caption{In spiking neural networks, the inference process generates its own output at each time step as time progresses. While the classification might be correct at certain moments, the overall prediction throughout the entire process can still be erroneous.}
	\label{inference_process}
\end{figure}

The non-differentiability of spikes necessitates the use of surrogate gradients to train Spiking Neural Networks (SNNs) \cite{wu2019direct,zheng2021going}. Within the framework that employs surrogate gradients for SNN training, the output typically integrates information across all time to make a final prediction \cite{fang2021deep,fang2021incorporating}. Concurrently, the time duration during the testing phase is usually constrained to be consistent with the training phase. To ensure uniformity in network architecture between training and prediction stages, an effective prediction usually requires a time duration equivalent to that of the training phase. Any reduction in time adversely affects model performance. Such constraints have become a key reason why SNNs are limited to extended time simulations, resulting in significant latencies. Additionally, these inflexible time restrictions and high latencies impede the advancement of SNNs in edge computing hardware. Furthermore, different devices might require varying durations to process SNNs, implying the need to retrain SNNs to adapt to new devices.

Another issue arising from this method of information aggregation is the inefficient use of temporal information. The aggregation process, which often is averaging, treats each time point as equally contributing to the final output. As illustrated in Figure \ref{inference_process}, we present an example to demonstrate this issue. During the training of the SNN, an image categorized as a 'panda' is input into the model. Outputs are obtained at six different time points, with correct outputs achieved at $T1$ and $T3$. However, after the aggregation process, the model predicts the category to be a 'bear'. The individual correct predictions are not weighted sufficiently to guide the model toward an overall accurate prediction. Some methods attempt to change the weights using attention mechanisms, but this usually results in increased computational and parameter overhead \cite{zhu2022tcja,yao2022attention}.
To address these two issues, we aim to develop a method for processing temporal information that amplifies the contribution of relevant data. This will allow the model to make more efficient use of information, while also freeing it from the constraint of requiring identical time steps across phases.

To rationally process information interactions across different moments, we see the aggregation of predictions from multiple time points as a form of model aggregation. Each time point is designated as a distinct sub-model, and the final output results from the aggregation of predictions from all these sub-models. The outputs from all moments collectively determine the final prediction after undergoing certain aggregation operations, such as weighting, selecting, voting, boosting, bagging\cite{sagi2018ensemble,polikar2012ensemble}. Throughout the computation, even though the model structure and weights remain constant, variations in input and the presence of membrane potentials mean that the membrane potential from a prior moment becomes part of the state and parameters for the subsequent moment. This gives each moment's output its unique information and significance. Even in the case of static image datasets, once they are encoded through SNNs, they still exhibit characteristics where different information is emphasized at different time points. 

To facilitate information transfer between models, we leverage the knowledge transfer capabilities of the knowledge distillation technique \cite{hinton2015distilling,mirzadeh2020improved}. The output at each moment encapsulates dark knowledge \cite{hinton2015distilling}, including class distribution probabilities that the model has learned. Fitting to this kind of knowledge can help the model approximate the capability of a teacher model. In this work, a teacher signal is derived from the combination of knowledge across different moments. As there's no external teacher model and knowledge transfer occurs between different moments of the same model, this approach can be considered as a form of temporal self-distillation.

In this study, we introduce the Temporal Knowledge Sharing (TKS) method. TKS employs a novel approach to harness temporal information, amplifying the contribution of relevant data to the final prediction by reorganizing temporal information. We have validated TKS on multiple static and neuromorphic datasets. The results demonstrate that the TKS's method of utilizing temporal information significantly enhances model performance. Additionally, tests on fine-grained datasets demonstrate the method's improvement in the model's ability to process detailed information. On the other hand, to verify that TKS can alleviate the requirement for consistent time, we also conducted tests with varying time steps. The results confirmed that TKS enables the model to substantially reduce performance degradation, even when there's a significant discrepancy between test and training time steps.

Our contributions can be summarized as follows:
\begin{enumerate}
\item We introduce a temporal self-distillation technique: Temporal Knowledge Sharing (TKS). By reorganizing temporal knowledge, TKS aims to address issues related to training time consistency and the inability to effectively leverage temporal information.
\item To validate the superiority of our model and its efficacy in harnessing temporal information, we conducted experiments on static datasets like CIFAR10, CIFAR100, and ImageNet-1K, as well as neuromorphic datasets such as dvs-CIFAR10 and NCALTECH101, and further tests on fine-grained classification datasets including CUB-200-2011, StanfordDogs, and StanfordCars demonstrated significant performance improvements in comparison to other state-of-the-art algorithms.
\item TKS effectively addresses the problem with time consistency. It enables high-performance inference using only a few number of time steps compared to the training phase.
\end{enumerate}

TKS provides a novel perspective on the relationship between information from multiple time points and their collective impact on the overall prediction. Furthermore, it addresses the problem with time consistency, significantly reducing the latency of SNNs, facilitating their convenient deployment on edge devices.

\section{Background}

\textbf{Knowledge Distillation} facilitates the transfer of knowledge from high-performance models to lower-performance models, aiming to achieve performance closer to the teacher model \cite{hinton2015distilling}. In the region of spiking neural networks, some employ additional knowledge from pre-trained artificial neural networks \cite{takuya2021training} or from large pre-trained SNNs \cite{kushawaha2021distilling} to guide the SNN training. These approaches necessitate extra training on larger networks, which incurs significant computational overhead.

Self-distillation \cite{yun2020regularizing,ji2021refine,kim2021self} offers an innovative approach by treating the model itself as the teacher. This process involves the model generating its own teacher signals for self-distillation. This eliminates the need for additional network training, substantially improving training efficiency and reducing computational costs.  Inspired by self-distillation, this paper presents a model that leverages its own knowledge at different time points to guide its training. This method obviates the need for additional network training and significantly reduces computational overhead.  The model proposed in this paper can be viewed as a temporal self-distillation model.

\textbf{Spiking Neural Network Training} Researchers are continuously exploring ways to enhance the performance of SNNs trained through backpropagation.  Some have sought to improve performance by modifying the structure of SNNs. For instance, inspired by biological structures, the LISNN model \cite{cheng2020lisnn} introduces lateral interactions in convolutional SNNs, boosting SNN performance and robustness. The BackEISNN model \cite{zhao2022backeisnn}, influenced by autapse mechanisms, integrates self-feedback connections to foster and expedite SNN training. Additionally, Other models derive insights from classical architectures in the realm of artificial neural networks, such as those based on ResNet \cite{fang2021deep} and Transformer \cite{zhang2022spiking}. Enhancements in the dynamic properties of spiking neurons, like learnable membrane time constants \cite{fang2021incorporating} and adaptive thresholds \cite{yin2021accurate}, can also amplify the representational capabilities of SNNs.

The potential of harnessing the temporal information within Spiking Neural Networks has also garnered attention. The TCJA-SNN model \cite{zhu2022tcja} incorporates an additional network to allocate attention mechanisms across temporal channels.  The AttentionSNN model \cite{yao2022attention} leverages temporal data by fine-tuning membrane potentials using attention weights. Other models like TET \cite{deng2022temporal} offer individualized supervision for each time point, while TEBN \cite{duan2022temporal} expands batch normalization to encompass the temporal dimension. Additionally, BPSTA \cite{shen2022backpropagation} pioneers the use of temporal residual connections to aid error backpropagation across time. However, there's limited research focusing on addressing constraints related to model training time.

\textbf{Ensemble Learning:} Ensemble learning is a technique that amalgamates multiple inducers  – often referred to as base learners – using aggregation methods. A base learner can be any type of machine learning algorithm. By combining these learners, errors made by an individual learner can be compensated for by others, leading to an ensemble model that typically outperforms any single base learner \cite{sagi2018ensemble}. Ensemble learning can prevent overfitting, avoid getting stuck in local minima, and achieve a more robust representation. For an ensemble to be effective, individual models should exhibit diversity, and the performance of each should be as high as possible. The efficacy of an aggregated model is proportional to the errors which are uncorrelated in sub-models   \cite{ali1995link}. Classic aggregation algorithms include AdaBoost \cite{schapire2013explaining,hastie2009multi}, Bagging \cite{breiman1996bagging}, Random Forest \cite{breiman2001random}, Random Subspace Methods \cite{ho1998random}, and Rotation Forest \cite{rodriguez2006rotation}.

In this paper, the existence of membrane potential in Spiking Neural Networks (SNNs) results in varying state parameters across different sub-models, ensuring diversity among the base learners. Integrating sub-models from different time points compensates for the errors of individual learners, leading to a final aggregated model with higher accuracy.

\begin{figure*}[!t]
	\centering
	\includegraphics[width=1.9\columnwidth]{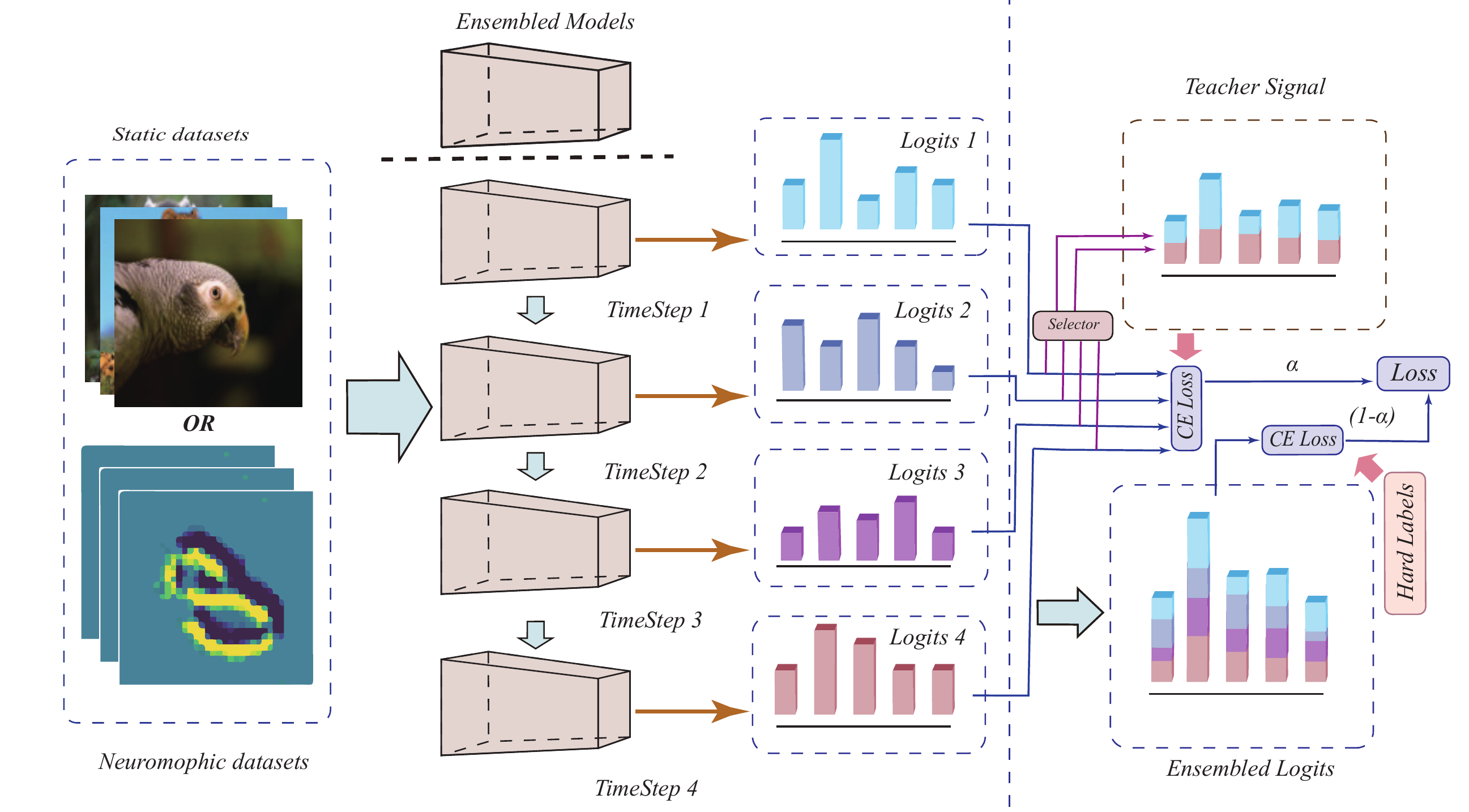} 
	\caption{The entire process involves temporal knowledge sharing. Spiking neural networks are perceived as temporal aggregation models, with each moment acting as a sub-model. TKS gathers knowledge from various time points and formulates an additional teacher signal to guide the network's training.}
	\label{backbone}
\end{figure*}

\section{Methodology}
\subsection{Spiking Neuron Model}
Spiking neurons serve as the fundamental computational units within spiking neural networks, utilizing differential equations to depict the dynamic behavior of biological neurons. Notably, the Leaky Integrate-and-Fire (LIF) neuron is the most prevalent neuron type in SNN training.  It characterizes the process where neurons accumulate membrane potential through inputs and emit a spike once they reach a threshold. These neurons grant spiking neural networks their unique ability for spatiotemporal processing. As shown in  Equation \ref{lif1}.
\begin{equation}
	\begin{aligned}
		\tau_m \frac{dV_t}{dt} &= -V_t + I_t \quad V_t \le V_{th} \\
		S_t &= 1, \quad V_t=V_{rest} \quad V_t \geq V_{th}
	\end{aligned}
	\label{lif1}
\end{equation}

By accumulating membrane potential over time, the neuron releases a spike after reaching the threshold $V_{th}$, and resets to the resting potential $V_{rest}$, which we set to 0 here. To facilitate computational simulation, we discretize Equation \ref{lif1} to obtain Equation \ref{lif2}. 

\begin{equation}
	V[t]=(1-\frac{1}{\tau})V[t-1](1-S[t-1])+\frac{1}{\tau}I[t]
	\label{lif2}
\end{equation}
$V[t]$ denotes the membrane potential at time $t$, $tau_m$ denotes the membrane potential constant, $I[t] = \sum_j w_{ij}S_j$ denotes the input current obtained by collecting pre-synaptic neuron spikes, and $S[t]$ denotes the spike at time $t$.

\subsection{Temporal Knowledge Sharing}
Owing to the dynamic properties of spiking neurons, the output of SNNs at each time point embodies unique knowledge and significance, as illustrated in the left half of Figure \ref{backbone}. We can view the SNN as a temporal aggregation model, where each time step represents a sub-model of the entire network, with each having its own output distribution. Variations in input and the presence of membrane potentials often result in diverse distributions.  Exchanging and sharing information across multiple time points can enhance the temporal processing capabilities of the SNN. This shared information is utilized during the model training and weight updating stages.

For a typical SNN, given a model $f$, each of its sub-models is represented as  $f_{t}$. Thus, for a given input  $x_{t}$, the corresponding output is $Q[t]^{out}=f_{t}(x_t,m_{(t-1)})$, where $m_{t}$ stands for the neuron membrane potential acquired at time $t$. The output of each sub-model is determined by the current input and the membrane potential from the previous time point. The final output consists of the average of all sub-model outputs.
 
\begin{equation}
	\begin{aligned}
		V[t]^{out} &= Softmax(Q[t]^{out})\\
		\mathbb{O} &= \mathbb{E}_t(V[t]^{out})
	\end{aligned}
	\label{out1}
\end{equation}

In this context,  $\mathbb{O}$ denotes the model's aggregated  output, while $V[t]^{out}$ signifies the output at the time $t$.

Knowledge transfer is achieved by using the logits' category distribution from each sub-model. This knowledge then provides synthesized teacher labels to guide other sub-models. As illustrated on the right side of Figure \ref{backbone}, every sub-model operates as a student. Meanwhile, any of these sub-models can also potentially serve as a teacher model, thereby contributing to the teacher signal. The objective for these sub-models is to learn by fitting to the teacher signal. To  measure the discrepancy between a sub-model's output and the teacher signal, a cross-entropy function is employed, as detailed in equation \ref{loss2}.

\begin{equation}
	L_{TKS} = -\mathbb{E}_t\left(\sum_{k=1}^n \mathbb{Z}_k log(V[t]^{out})\right)
	\label{loss2}
\end{equation}

Where $\mathbb{Z}$ denotes the teacher signal, and the discrepancy between the sub-model and the teacher signal constitutes the error $L_{TKS}$. The teacher signal is derived by adaptively aggregated various sub-models based on their respective inputs and outputs. Several strategies for model aggregation exist \cite{sagi2018ensemble,polikar2012ensemble}, including but not limited to averaging, selection, weighting, boosting, and bagging. To ensure stability during training in this study, we choose sub-models whose output distributions closely align with the labels to form the teacher models.

Additionally, as demonstrated in Equation \ref{loss3}, we introduced a temperature parameter \cite{hinton2015distilling} into the softmax operation of the teacher signal to ensure a smoother fitting process. A higher $\tau$ value corresponds to increased smoothness.

\begin{equation}
	\mathbb{Z}_{ki}=\frac{exp(Q_{i}[k]^{out} /\tau)}{\sum_j exp(Q_{j }[k]^{out}/\tau)}
	\label{loss3}
\end{equation}
$\tau$ signifies the temperature parameter, and $Q_{j }[k]^{out}$ corresponds to the output from the selected k-th sub-model.

\begin{algorithm}[h]
	\caption{Temporal Knowledge Sharing For Training Deep Spiking Neural Networks}
	\label{alg:Framwork}
	\begin{algorithmic}[1] 
		\REQUIRE ~~\\ 
		Training dataset D = $\{(x_i, y_i)\}_{i=1}^N$, SNN model $f$, time Length $T$, 
		\ENSURE ~~\\ 
		A high-performance, low-latency and stable spiking neural network model;
		\FOR{each mini-batch training data $D_i = \{x_i, y_i\}$}
		\STATE Compute each sub-model outputs $V[t]^{out} $
		\STATE Compute the SNN output $\mathbb{O}$ as shown in Equation \ref{out1}
		\STATE Aggregate the teacher signal $\mathbb{Z}$ 
		\STATE Compute the TKS loss as shown in Equation \ref{loss2}
		\STATE Compute the CE loss as shown in Equation \ref{loss1}
		\STATE Compute the final loss using Equation \ref{loss_all}
		\STATE Adjust the synaptic weights through the spatial temporal backpropagation with surrogate gradients 
		\ENDFOR
	\end{algorithmic}
\end{algorithm}

The model's loss function constitutes two components, as depicted in Equation \ref{loss_all}. It consists of the standard cross-entropy loss with respect to the labels, as described in Equation \ref{loss1}, as well as a fitting loss that measures the discrepancy between the sub-model and the teacher signal, as illustrated in Equation \ref{loss2}.

\begin{equation}
	L_{CE} = -\sum_{i=1}^n y_i log(\sum_{t}{V_{i}[t]^{out}})
	\label{loss1}
\end{equation}

\begin{equation}
	L_{final}=
	(1-\alpha)L_{CE}+\alpha \tau^2 L_{TKS}
	\label{loss_all}
\end{equation}

 For a sub-model, its loss is composed of two components: the discrepancy between the aggregated output and the labels, and the difference between its own output and the teacher signal, as depicted in Equation \ref{loss4}. The sub-model undertakes learning and shares knowledge from other sub-models through the $L(V[k]^{out} ||\mathbb{Z})$ portion.

 \begin{equation}
 	L_{sub}[t]=
 	\frac{(1-\alpha)L_{CE}}{T}+\alpha \tau^2 L(V[k]^{out} ||\mathbb{Z})
 	\label{loss4}
 \end{equation}
 
 The coefficient $\alpha$ serves as a balance factor, adjusting the equilibrium between the loss associated with predictions and the actual labels and the loss between predictions and the teacher signal. A larger value of $\alpha$ indicates that the teacher signal plays a more crucial role in the model's weight adjustments. Since the model cannot provide a highly accurate supervisory signal in the early stages of training, labels are necessary for primary guidance. In this context, we adopt a progressive ascent strategy: initializing $\alpha$ with a small value and allowing it to linearly increase throughout the training. Based on experience, the initial value of $\alpha$ is set to 0 and gradually increases during training, reaching its peak in the final training epoch.

\begin{table}[h]
	\caption{Accuracy of static datasets: CIFAR10, CIFAR100, ImageNet-1K. Where * denotes data augmentation, The best performing model is indicated as boldface.}
	\centering
	\scalebox{1.0}{
		
		\setlength{\tabcolsep}{1.0mm} {
			\begin{tabular}{lllrr}
				\toprule[1.5pt]
				Model&Method & Architecture&T&Accuracy(\%) \\
				\midrule
				
				CIFAR10 &&&& \\
				
				\cite{rathi2020enabling} & \makecell[ll]{Hybrid\\ training }    &  ResNet-20 &250&92.22\\
				\cite{han2020rmp}& ANN2SNN    & ResNet-20& 2048&91.36\\
				\cite{deng2021optimal}&ANN2SNN&ResNet-20&128&93.56\\
				\cite{wu2018spatio}     & STBP     & CIFARNet &12&89.83 \\
				\cite{li2021differentiable} & DSpike     & ResNet-18 &4&93.66\\
				\cite{zhang2020temporal} & TSSL-BP     & CIFARNet &5&91.41\\
				\cite{zheng2021going}     & STBP-tdBN      & ResNet-19&4 &92.92\\
				\cite{deng2022temporal} & TET*     & ResNet-19&4&94.44 \\
				\cite{duan2022temporal}&TEBN*&ResNet-19&4&95.58\\
				\cmidrule(r){1-5}
				Ours		  & TKS  & ResNet-19     & 4&\textbf{95.30} \\
				Ours  & TKS*      & ResNet-19  & 4&\textbf{96.35}\\
				Ours & TKS*	 	&SEW-ResNet-19 &4&\textbf{96.76}\\
				\midrule

				CIFAR100 &&&&\\
				
				\cite{rathi2020enabling} &  \makecell[ll]{Hybrid\\ training }    &  VGG11 &125&67.87\\
				\cite{rathi2020diet}  & Diet-SNN     & ResNet-20&5&64.07\\
				\cite{han2020rmp}& ANN2SNN     & ResNet-20& 2048&67.82\\
				\cite{li2021differentiable} & DSpike     & ResNet-18 &4&73.35\\
				\cite{zheng2021going}     &  STBP-tdBN     & ResNet-19&4&70.86\\
				\cite{deng2022temporal} & TET*     & ResNet-19&4&74.47 \\
				\cite{duan2022temporal}&TEBN*&ResNet-19&4&78.71\\
				\cmidrule(r){1-5}
				Ours & TKS      & ResNet-19  &4&\textbf{76.20}\\
				Ours & TKS*     & ResNet-19 &4&\textbf{79.89}\\
				Ours & TKS*		&SEW-ResNet-19 &4&\textbf{80.67}\\
				\midrule
				ImageNet-1K &&&&	\\
				
				\cite{rathi2020enabling} &  Hybrid  training     &  ResNet-34 &250&61.48\\
				\cite{sengupta2019going}     &  SPIKE-NORM      & ResNet-34&2500&69.96 \\
				\cite{zheng2021going}     & STBP-tdBN     &   Spiking-ResNet-34 &6&63.72\\
				\cite{fang2021deep} & SEW ResNet   &SEW-ResNet-34&4&67.04\\
				\cite{fang2021deep} & SEW ResNet   &SEW-ResNet-50&4&67.78\\ 
				\cite{deng2022temporal} & TET     & SEW-ResNet-34&4&68.00 \\
				\cite{duan2022temporal} & TEBN     & SEW-ResNet-34&4&68.28 \\ 
				\cmidrule(r){1-5}
				Ours & TKS     & SEW-ResNet-34 &4&\textbf{69.60}\\
				Ours & TKS     & SEW-ResNet-50 &4&\textbf{70.70}\\
				
				\bottomrule
			\end{tabular}
	}}

	\label{statictab}
\end{table}

\section{Experiment}
TKS efficiently leverages temporal information by interchanging knowledge from sub-models across different time intervals. Additionally, TKS overcomes the limitations of temporal consistency. To demonstrate the effectiveness of TKS in addressing these issues, we conducted extensive experiments. Tests were carried out on multiple datasets for classification tasks, and the model was evaluated under conditions where the test and training time steps were different.
To showcase the superiority of the proposed algorithm, validation was conducted on commonly used static datasets such as CIFAR-10, CIFAR-100, and ImageNet-1K, as well as neuromophic datasets like DVS-CIFAR10 and NCALTECH101. In these experiments, we employed the surrogate gradient function from \cite{liu2018bi}. For all datasets except ImageNet-1K, the AdamW optimizer \cite{loshchilov2017decoupled} was used for model optimization with a weight decay set at 0.01. For ImageNet-1K, the Adam optimizer \cite{kingma2014adam} was utilized, and the weight decay was set to zero. The batch size for all experiments was set to 128, and a cosine annealing strategy was adopted to control the learning rate. The spiking neuron threshold was set at 0.5, and $\tau_m$  was set at 2.

\begin{table}[t!]
	\centering
	\caption{Accuracy of neuromophic datasets: DVS-CIFAR10, NCALTECH101. The best result are shown in boldface.}
	\scalebox{1.0}{
		
		\setlength{\tabcolsep}{1.5mm} {
			\begin{tabular}{lllrr}
				\toprule[1.5pt]
				Method  & Model&Architecture&T & Accuracy(\%) \\
				\midrule			
				 DVS-CIFAR10&&&&\\
				 \cite{zheng2021going}     &  STBP-tdBN      &  \makecell[l]{ResNet-19}&10&67.8\\
				\cite{kugele2020efficient}  &  Streaming Rollout      & DenseNet&10&66.8 \\
				\cite{wu2021liaf} &  Conv3D     & LIAF-Net&10&71.7 \\
				\cite{wu2021liaf} &  LIAF     & LIAF-Net&10&70.4   \\
								\cite{li2021differentiable} & DSpike     & ResNet-18 &10&75.4\\ 
				\cite{fang2021incorporating} & PLIF & CNN6&20&74.8\\
				\cite{zhu2022tcja}     &  TCJA-SNN     & VGG-SNN&10&80.7 \\
				\cite{deng2022temporal} & TET     & VGG-SNN&10&83.2 \\
				\cite{duan2022temporal}&TEBN&VGG-SNN&10&84.9\\
			\cmidrule(r){1-5}
				Ours & TKS     & VGG-SNN &10&\textbf{85.3}\\
				\midrule
				NCALTECH101&&&&\\
				\cite{zhu2022tcja}     & TCJA-SNN     & TCJAnet&14&78.5 \\
								\cite{shen2022eventmix}  & EventMixer&    ResNet-18 & 10&79.5 \\
				\cite{li2022neuromorphic} & NDA     & ResNet-19&10&78.6 \\
				\cite{sironi2018hats} & HATS     & - &-&64.2\\
				\cite{ramesh2019dart} & DART     & -&-&66.8 \\ 
								\cmidrule(r){1-5}
				Ours & TKS     & VGG-SNN &10&\textbf{84.1}\\
				\bottomrule
	\end{tabular}}}

	\label{neuromophictab}
\end{table}

\subsection{Static Datasets}

Even with identical inputs, the presence of membrane potentials and spiking neurons means that neuron outputs have different information and meanings at different times. Consequently, we explored whether temporal knowledge sharing could enhance the model's performance on static datasets.

We validated the TKS algorithm on the CIFAR-10, CIFAR-100, and ImageNet-1k datasets and compared it with existing methods. For a fair comparison on CIFAR-10 and CIFAR-100 \cite{krizhevsky2009learning}, we employed the ResNet-19 \cite{zheng2021going} and SEW-ResNet-19 structures \cite{deng2022temporal}. The SEW-ResNet-19 improves upon the original ResNet, allowing for more efficient residual learning in SNNs. In a comparable fashion, we used the SEW-ResNet-34 and SEW-ResNet-50 structures for the ImageNet-1k dataset. For CIFAR-10 and CIFAR-100, the temperature parameter $\tau$ was set to 3, while for the ImageNet-1k dataset, $\tau$ was set to 1. 
Throughout the experiments, $\alpha$ increased from an initial value of 0 to 0.7.

As shown in Table \ref{statictab}, TKS method outperformed other deep SNN algorithms on all datasets. Compared to the TET and TEBN algorithms, TKS showed significant improvements on both CIFAR-10 and CIFAR-100. Additionally, the SEW-ResNet-19 structure further enhanced model performance. Notably, for the ImageNet-1k dataset and under the same network structure of SEW-ResNet-34, our performance exceeded theirs by 1.3\% and 1.6\%, respectively.

The aforementioned results indicate that the information sharing approach of the TKS method indeed assists the model in better processing information.

\subsection{Neuromorphic Datasets}
Neuromorphic datasets are typically recorded using event-based cameras \cite{gallego2020event}, where the recorded data consists of events. Each event signifies a change in pixel brightness. Neuromorphic datasets effectively capture object changes and usually exhibit strong temporal dependencies. Hence, we investigated the performance of temporal knowledge sharing on neuromorphic datasets. Experiments were conducted on the DVS-CIFAR10 \cite{li2017cifar10} and NCALTECH101 \cite{orchard2015converting} datasets, both of which are derived from their respective original datasets using DVS camera. For a fair comparison, we used the VGG-SNN model \cite{deng2022temporal} as the base architecture. The temperature was set to 5, and $\alpha$ increased from an initial value of 0 to 0.3.

As shown in Table \ref{neuromophictab}, compared to the current best-performing algorithms, TKS achieved the highest accuracy, obtaining 85.3\% on DVS-CIFAR10 and 84.1\% on NCALTECH101.

The aforementioned results suggest that the TKS method indeed enhances the model's ability to process temporal information effectively on neuromorphic data.


\subsection{Addressing Temporal Consistency}

\begin{figure}[t]
	\centering
	\subfloat[\label{timefig:a}]{\includegraphics[width=0.99\columnwidth]{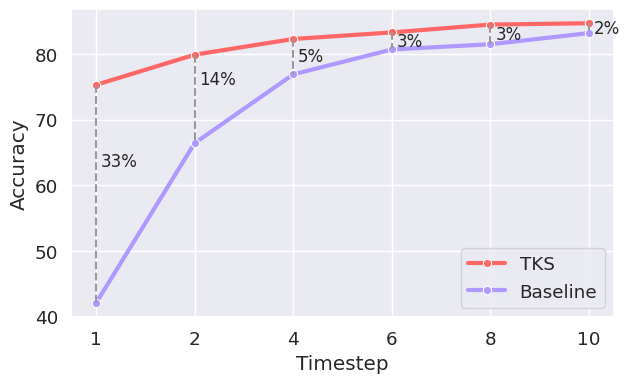}}\\
	\subfloat[\label{timefig:b}]{\includegraphics[width=0.99\columnwidth]{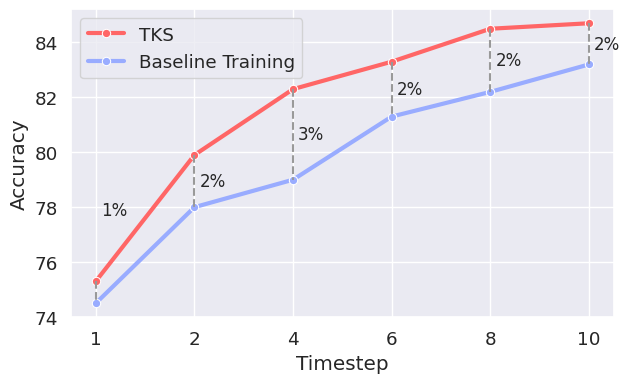} }
	
	\caption{(a) The accuracy across different test time steps on the DVS-CIFAR10 dataset, with the model being trained at a time step of 10. (b)  The accuracy across different time steps, with the model being trained at related time step.}
	\label{timefig}
\end{figure}

Beyond the performance of Spiking Neural Networks (SNNs), another challenge hampering their development is latency. SNNs often require extended time steps to produce accurate results. Moreover, the time step during the testing phase usually needs to match that of the training phase. Additionally, latency can vary when dealing with different edge devices. This implies that when changing devices, models need to be retrained with different time steps, severely reducing efficiency. We aspire for SNNs to achieve higher performance with larger time steps during training and to deliver near-lossless performance during the inference phase with smaller time steps. To validate TKS's capability in addressing the issue of temporal consistency, we conducted experiments.

As depicted in Figure \ref{timefig}(a), the performance of the VGG-SNN model trained on the DVS-CIFAR10 dataset is showcased. During the training phase, we set the timestep to 10. In the testing phase, we evaluated performance across different time steps. When the testing time step was inconsistent with the training time step, the SNN model's performance dramatically declined. For the baseline SNN model, with a time step of 1 during the testing phase, network performance plummeted by nearly 33\%. Our TKS still managed to achieve an accuracy of around 75\%. As the time step increased, the performance gap between the two narrowed.

Additionally, to validate the robustness of the TKS method across various time steps, we trained and tested the SNN model using multiple distinct time steps, keeping the training and testing steps consistent. For instance, when the timestep is set to 1, both training and testing utilize a step size of 1. As can be seen in Figure \ref{timefig}(b), our TKS model outperforms others across different timesteps. The TKS model showcases strong generality across time steps, offering considerable convenience for the development of SNNs.

\begin{figure}[!htp] 
	\subfloat[\label{singletimefig:a}]{\includegraphics[width=0.99\columnwidth]{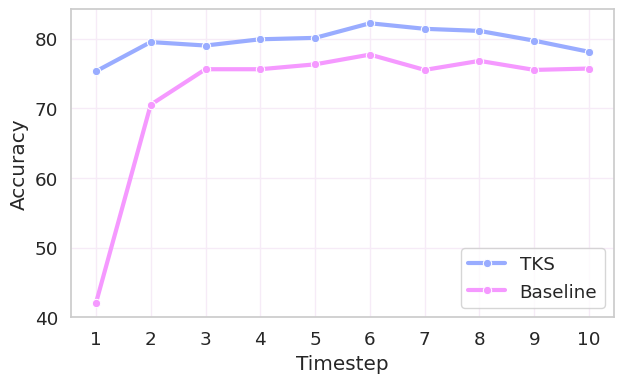}}\\
	\subfloat[\label{singletimefig:b}]{\includegraphics[width=0.99\columnwidth]{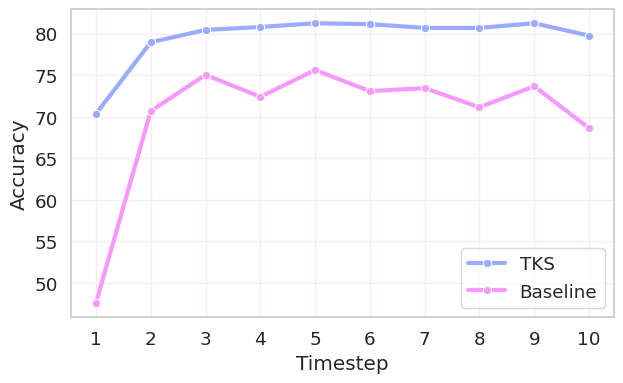}}
	
	\caption{Accuracy of sub-models at each time point on the DVS-CIFAR10 (top) and NCALTECH101 (bottom) datasets, with training conducted at a time step of 10. }
	\label{singletimefig}
\end{figure}

To investigate how TKS assists SNNs in addressing the issue of temporal consistency, we separately examined the impact of each sub-model on the aggregated model, i.e., the corresponding accuracy at each time point. We visualized the accuracy of each point for both TKS and the baseline. The performance was verified on DVS-CIFAR10 and NCALTECH101, with models trained at a timestep of 10. As depicted in Figure \ref{singletimefig}, for TKS, the output performance at each instance is superior to the baseline model. Interestingly, we observed that the performance of the first sub-model in the baseline is typically the lowest and significantly affects the aggregated model's performance. We speculate that this is due to the neuron membrane potential resetting at the first instance, causing its distribution to differ considerably from subsequent sub-models. It's evident that TKS significantly enhances the accuracy of the sub-model at each instance and greatly mitigates the performance dip in the initial moment.

Thus, TKS aids the model in achieving better performance at each time instance. Through knowledge sharing, it empowers the model with enhanced capabilities to process data individually at every moment.

\begin{figure*}[t]
	\centering
	\subfloat[\label{tsnefig:a}]{\includegraphics[width=0.49\columnwidth]{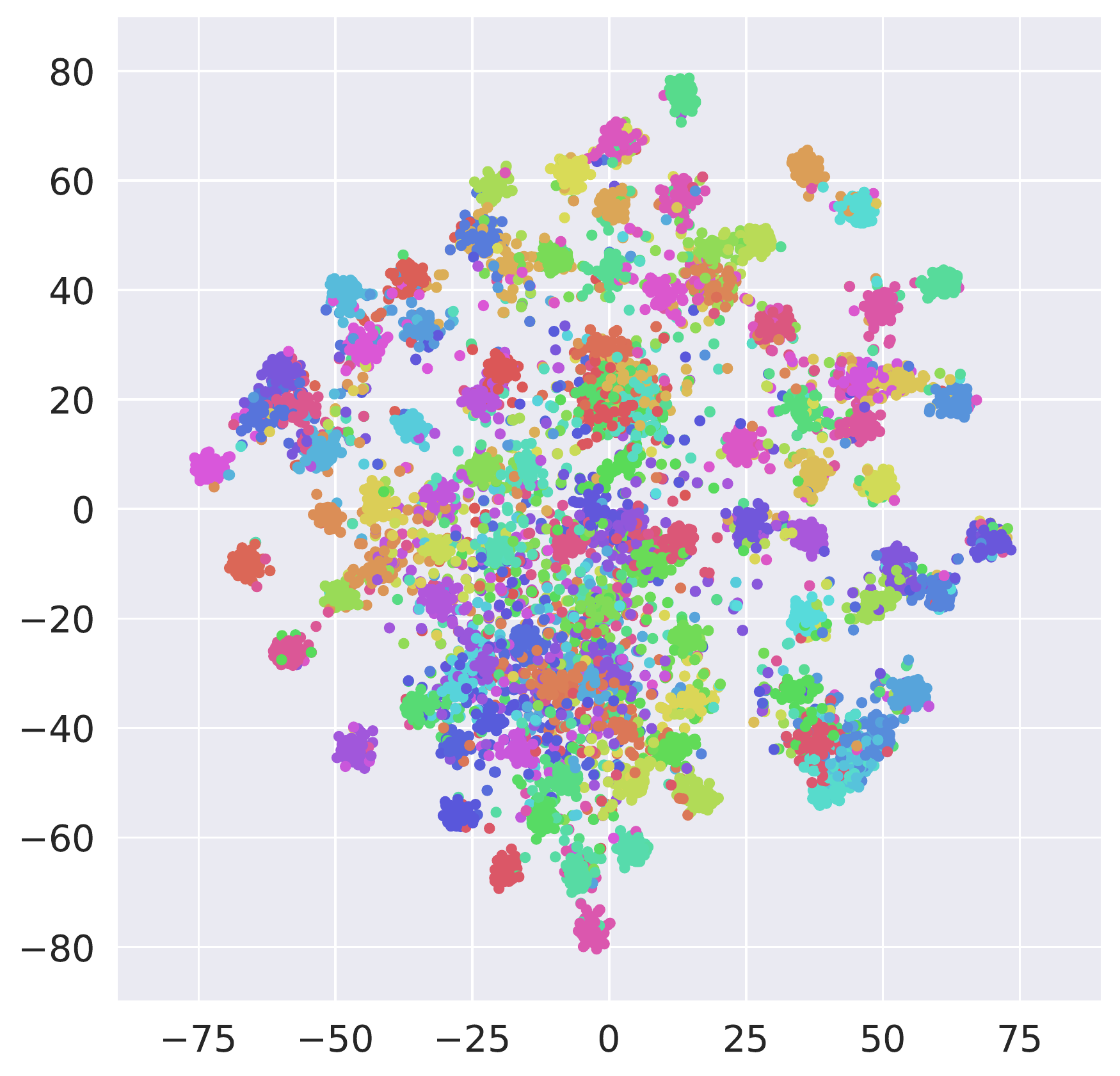}}
	\subfloat[\label{tsnefig:b}]{\includegraphics[width=0.49\columnwidth]{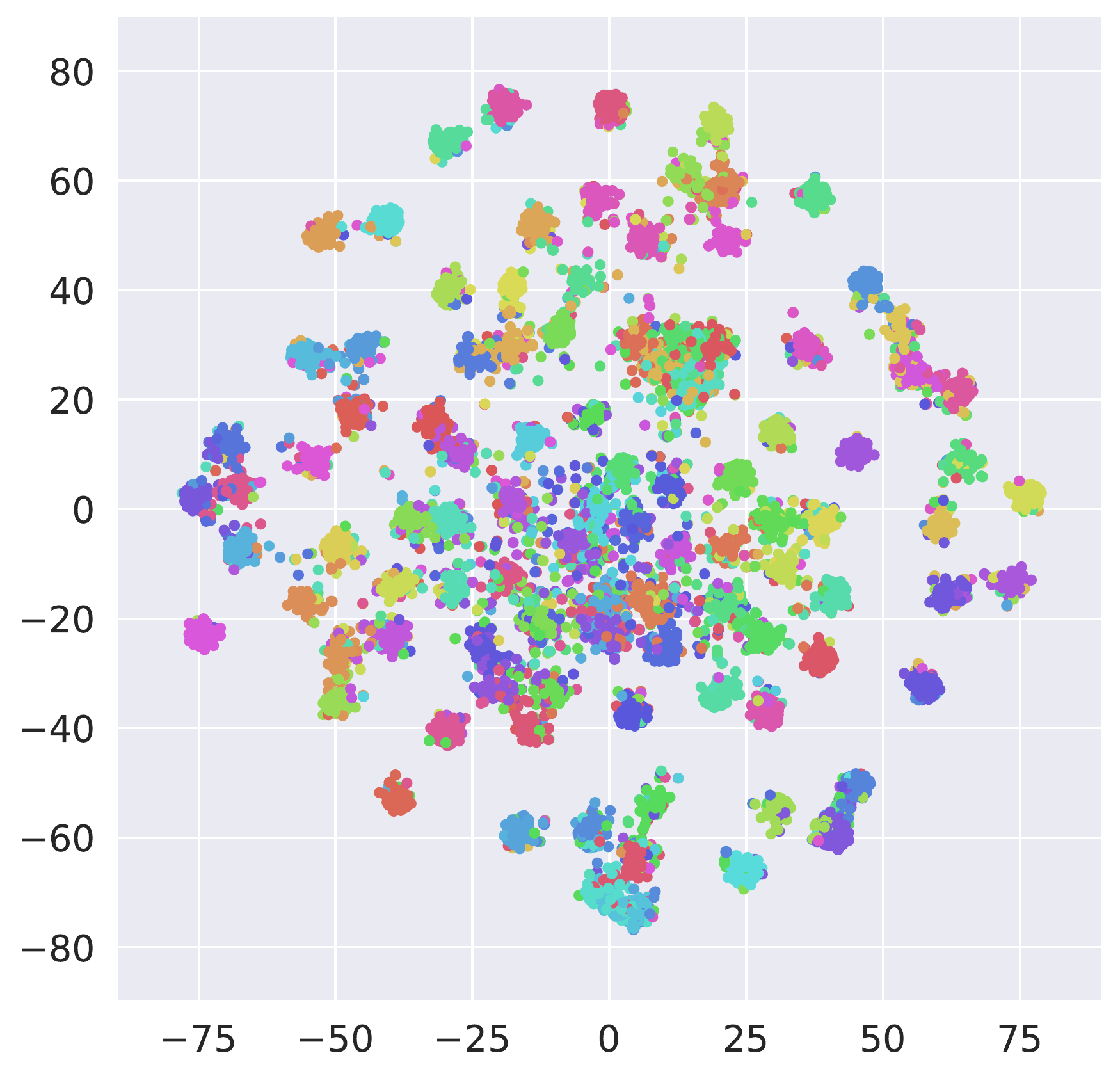}}
	\subfloat[\label{tsnefig:c}]{\includegraphics[width=0.49\columnwidth]{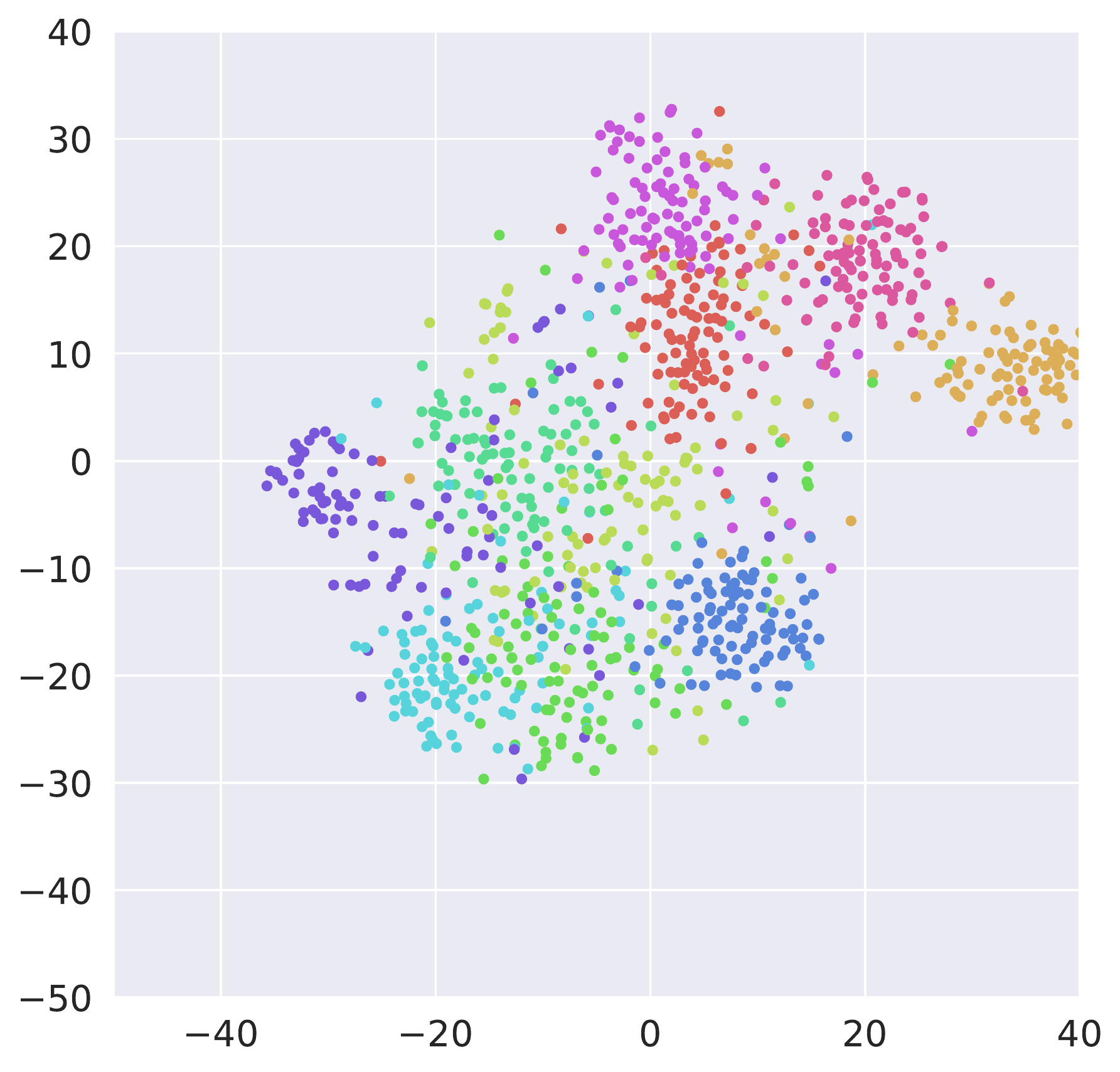}}
	\subfloat[\label{tsnefig:d}]{\includegraphics[width=0.49\columnwidth]{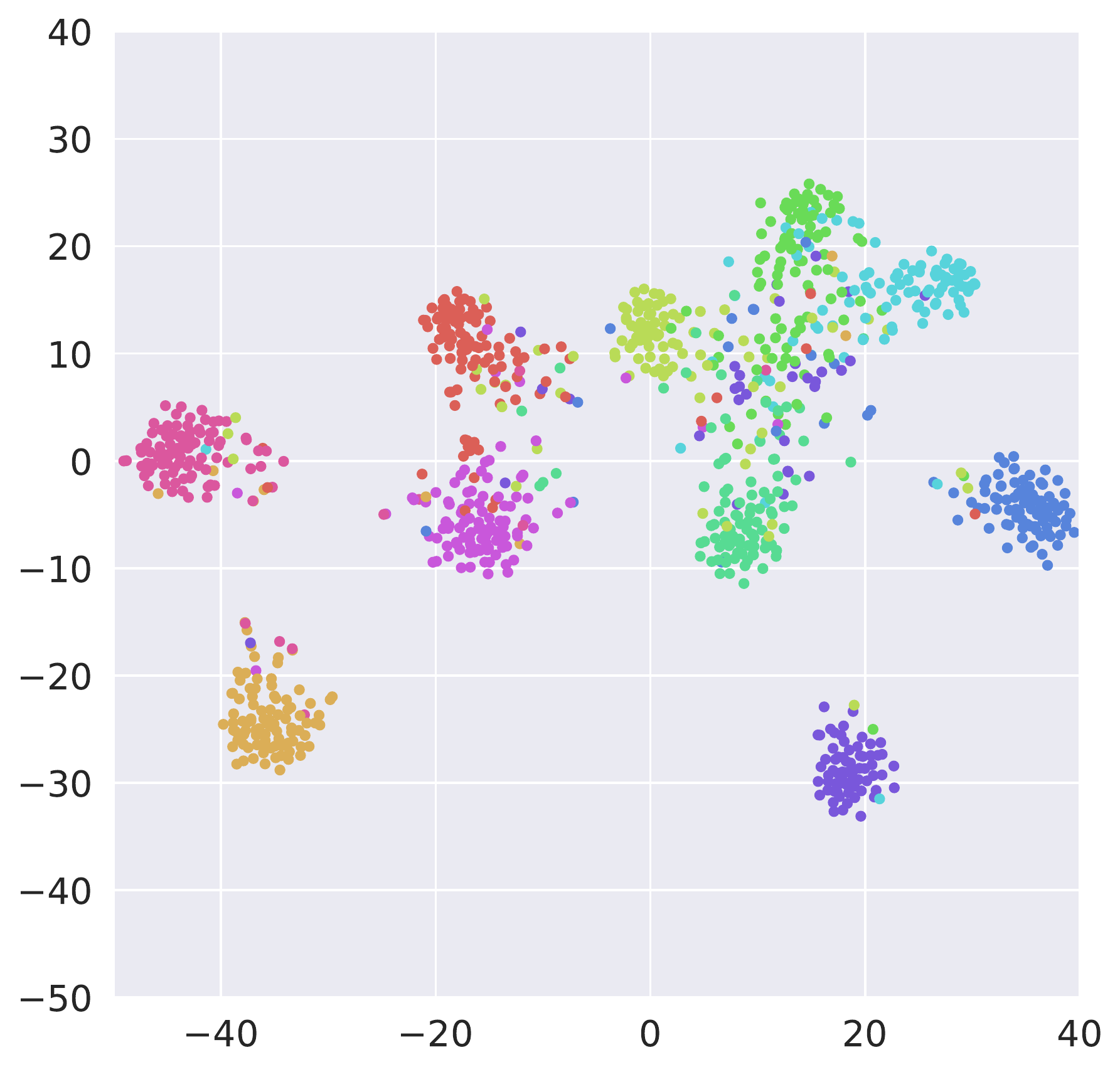}}
	\caption{Visualization results from t-SNE:
		(a) Baseline results for CIFAR100.
		(b) TKS results for CIFAR100.
		(c) Baseline results for DVS-CIFAR10.
		(d) TKS results for DVS-CIFAR10.}
	\label{tsnefig}
\end{figure*}

Also, we verify the performance on DVS-CIFAR10 as well as NCALTECH101 using the output of SNN at each single moment. The models are trained in timestep 10. As shown in Fig.~\ref{singletimefig}, for TKS, the performance of the output at each moment is higher than that of the baseline model. The teacher signal helps the model to achieve better performance at each moment. Thus the model at each moment acquires the ability to process the data separately.

\section{Discussion}
In this section, we will carry out an ablation study. Within our experimental parameters, we compare TKS against several methodologies, highlighting its superior performance. We further illustrate the efficacy of TKS in processing fine-grained information by examining its role in fine-grained classification tasks. Moreover, we probe the influence of different categories on the end results through experimentation. Lastly, we employ the t-SNE method to visualize feature embeddings.
\subsection{Evaluation metrics}
\textbf{Accuracy:} The Top-1 accuracy reflects the model's classification accuracy on the corresponding dataset. Here, "Top-1" denotes the accuracy of the category with the highest predicted probability.

\textbf{AURC (Area Under the Risk Coverage Curve) \cite{geifman2018bias}:} AURC quantifies the area under the Risk Coverage (RC) curve during the convergence process. It describes the quality of probabilistic predictions from a confidence estimation perspective. This metric assesses how well correct and incorrect predictions can be distinguished from one another.

%
\begin{table}[h]
	\caption{Comparison of the multiple methods implemented on our Baseline. The best performing model is indicated as boldface.}
	\centering
	\scalebox{1.0}{
		
		\setlength{\tabcolsep}{2.5mm} { 
			\begin{tabular}{l|llr|r}
				
				\toprule[1.5pt]
				Dataset  &Model&Method& \makecell{Top1\\Acc(\%)}  & \makecell{AURC\\ (x$10^{3}$) }\\
				\midrule
				
				\multirow{8}{*}{CIFAR10}			& \multirow{4}{*}{ResNet19}   &Baseline & 95.76     & 4.10  	\\
				& 			 				  &LS 		& 95.99     & 6.60	\\
				& 							  &TET		& 96.14     & 3.64	\\
				& 							  &TKS 	& \textbf{96.35}      & \textbf{2.76}	\\ 	
				& \multirow{4}{*}{\makecell{SEW-\\ResNet-19}}    &Baseline & 96.54  & 3.50  	\\
				& 							  &LS 		& 96.54    & 8.44	\\
				& 							  &TET		& 96.41     & 3.59	\\
				& 							  &TKS 	& \textbf{96.76}   &	\textbf{2.75}	\\			     			
				\midrule 
				\multirow{8}{*}{CIFAR100}			& \multirow{4}{*}{ResNet19}   &Baseline & 76.78        & 78.27  \\
				& 			 				  &LS 		& 76.77 	  &71.96						\\
				& 							  &TET		&78.90  &65.28						\\
				& 							  &TKS 	& \textbf{79.89}   &\textbf{54.07}	\\
				& \multirow{4}{*}{\makecell{SEW-\\ResNet-19}}    &Baseline & 78.35      & 67.55  \\
				& 							  &LS 		& 79.39 &63.43	\\
				& 							  &TET		& 79.53 	 	 &58.93 							\\
				& 							  &TKS 	& \textbf{80.67}  &	\textbf{54.56}						\\
				
				\midrule
				\multirow{4}{*}{ImageNet-1K}	& \multirow{3}{*}{\makecell{SEW-\\ResNet-34}}    &Baseline & 68.61   & -  \\
				& 							  &TET 		& 68.68 	 	 &-			\\
				& 							  &TKS		& 69.60  &-
				\\
				
				& 					\makecell{SEW-\\ResNet-50}	  &TKS 	&  \textbf{	70.70} 		 &-				\\
				
				\midrule
				\midrule 		
				\multirow{4}{*}{\makecell{DVS-\\CIFAR10}}	& \multirow{4}{*}{VGG-SNN}    &Baseline & 83.20       & 46.14  \\
				& 							  &LS 		& 84.00   	 &40.20						\\
				& 							  &TET		& 84.70 	 	 &43.42			\\
				& 							  &TKS 	&\textbf{85.30}   	&\textbf{40.18}				\\
				\midrule 		  
				
				\multirow{4}{*}{\makecell{NCAL-\\TECH101}}	& \multirow{4}{*}{VGG-SNN}    &Baseline & 78.26     & 52.03  \\
				& 							  &LS 		& 82.76     &	35.21					\\
				& 							  &TET		& 81.72  &36.89							\\
				& 							  &TKS 	& \textbf{84.10}   &\textbf{30.00}							\\
				
				\bottomrule[1.5pt]
	\end{tabular}}}
	
	\label{ablationttab}
\end{table}

\subsection{Ablation analysis}
To demonstrate the efficacy of the TKS algorithm, ablation experiments were conducted on the CIFAR10, CIFAR100, ImageNet-1K, DVS-CIFAR10, and NCALTECH101 datasets.For a fair comparison with existing methods, we benchmarked the LS \cite{szegedy2016rethinking} and TET \cite{deng2022temporal} methods against ours.

The LS (Label Smoothing) approach is a regularization technique, which can also be seen as passing teacher information that follows a uniform distribution \cite{muller2019does,yuan2020revisiting}. It's a common strategy to prevent models from becoming overly confident. Including LS in our comparisons ensures that the teacher signal provided by TKS genuinely conveys valuable information, rather than merely serving as a regularization term.

The TET algorithm applies individual supervision at each time point, aiming to exploit temporal information. By integrating TET in our comparisons, we ensure the same setting across individual time-point supervisions, while verifying that the teacher signal indeed imparts more information than the mere label signal.

As shown in Table \ref{ablationttab}, TKS achieved the highest performance when compared with the baseline, TET, and LS methods. This underscores the capability of TKS to effectively harness temporal information to enhance model performance. Moreover, to demonstrate the superiority of TKS, we evaluated the model from a confidence perspective. We computed the AURC for each model. As evident from Table \ref{ablationttab}, a lower AURC indicates higher quality in probabilistic predictions.

\subsection{Fine-grained Classification Task}
Fine-grained datasets typically exhibit similar features among categories, making them challenging to distinguish \cite{yang2018learning}. Moreover, the number of samples per category is limited. Fine-grained tasks are often employed to gauge a model's feature extraction capabilities. By exploring TKS's performance on fine-grained tasks, we demonstrate that TKS enhances SNN's feature extraction prowess by leveraging temporal knowledge.
\begin{table}
	\caption{Accuracy of fine-grained task datasets: CUB-200-2011, Stanford Dogs, Stanford Cars. The best result are shown in boldface. It demonstrates both top1 Acc and top5 Acc.}
	\centering
	\setlength{\tabcolsep}{1.5mm} {
		\begin{tabular}{llrr|rr|rr }
			\toprule[1.5pt]
			\multirow{2}{*}{ Model}&\multirow{2}{*}{ Method}  & \multicolumn{2}{c}{CUB-200-2011}&\multicolumn{2}{c}{StanfordDogs}&\multicolumn{2}{c}{StanfordCars}    \\
			\cmidrule(r){3-8}
			&& top1     &top5		&  top1   &top5&  top1   &top5 \\
			\midrule
			\multirow{3}{*}{\makecell{SEW-\\ResNet-18}}&Baseline    &37.94     & 63.55 &51.54&78.10&68.59 &88.96\\ 
			&TET  		& 40.49     & 65.53 &52.14&78.69&68.52&88.56\\
			&TKS 		& \textbf{42.82}     & \textbf{66.59} &\textbf{54.03}&\textbf{79.76}&\textbf{71.22}&\textbf{89.70}\\
			\multirow{3}{*}{\makecell{SEW-\\ResNet-34}}	&Baseline    & 46.77     & 72.35 &56.79&\textbf{83.57}&76.04&92.71\\ 
			&TET  		& 47.38     & 71.06 &57.06&81.98&72.94&92.81\\
			&TKS 		& \textbf{51.71}    & \textbf{75.28} &\textbf{59.06}& 82.82 &\textbf{76.94}&\textbf{92.95}\\		
			\bottomrule
		\end{tabular}
	}
	\label{fiegrainedtab}

\end{table}
We conducted experiments on the CUB-200-2011 \cite{wah2011caltech}, StanfordDogs \cite{wah2011caltech}, and StanfordCars \cite{krause2013collecting} datasets, and compared TKS against baseline and TET models. Both Top1 and Top5 accuracy metrics were presented. For the fine-grained tasks, the temperature parameter $\tau$ was set to 3, while $\alpha$ grew from an initial value of 0 to 0.7.

As illustrated in Table \ref{fiegrainedtab}, our TKS surpasses the baseline in terms of the Top1 metric across all datasets and even outperforms the TET algorithm. Under two experimental structures, sew-resnet18 and sew-resnet34, TKS consistently exhibited performance enhancements, underlining its robustness. By adopting the distribution information of output categories as the teacher signal, TKS assists the model in amplifying its classification capabilities, especially for categories with closely related features.
\begin{figure}[b]
	\includegraphics[width=0.75\columnwidth]{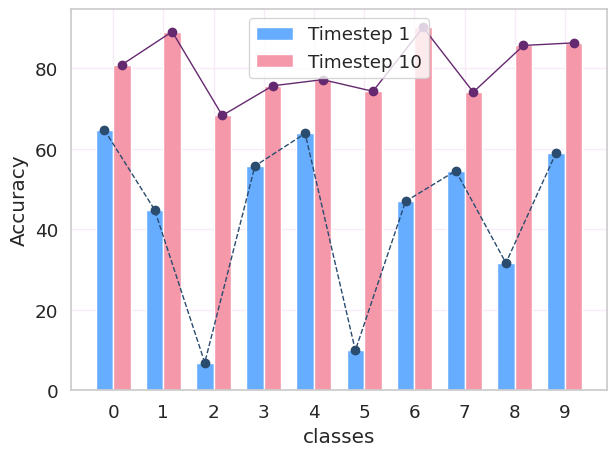}
	\centering
	\caption{The accuracy across different classes on the DVS-CIFAR10 dataset at test time step of 1, with the model being trained at a time step of 10.}
	\label{classesfig}
\end{figure}
\subsection{Impact of Timestep on Different Categories}
To further investigate the significant decrease in accuracy during the test phase,  we visualized the accuracy for each category at the first timestep. We trained the SNN model using the DVS-CIFAR10 dataset. Comparing the results between the baseline and TKS methods,  we observed that for several categories, the baseline model experienced significant drops in accuracy, which could be the reason behind the overall decline in performance. On the other hand, the TKS approach did not exhibit such drastic drops across categories.

\subsection{Feature Visualization}
We employed the t-SNE visualization technique \cite{van2008visualizing} to depict the feature embeddings from the penultimate layer of our model. This visualization approach sheds light on TKS's capability to enhance the model's feature extraction. We present results for both TKS and the baseline model on CIFAR100 using ResNet-19, and on DVS-CIFAR10 using VGG-SNN, as depicted in Figure \ref{tsnefig}. Compared to the baseline, TKS demonstrates a more effective separation within categories both in static and neuromorphic image data. The teacher signals from TKS work towards reshaping the distribution of sample features, amplifying the separations between similar categories, and enabling the classifier to differentiate more effortlessly.

\section{Conclusion}
The efficiency and latency of SNNs have long been significant bottlenecks hindering their progression. This study sheds light on the temporal consistency issues within spiking neural networks, as well as the inefficient utilization of temporal information. Addressing these concerns, we introduce a novel method named Temporal Knowledge Sharing (TKS). This approach perceives SNNs as temporal aggregation models and endeavors to enhance model training by facilitating knowledge interchange across distinct time points. TKS facilitates model training by enabling knowledge transfer across different time steps. This strategy can be regarded as a form of temporal self-distillation.

Our experiments, conducted on both static datasets and neuromorphic datasets, underscore the efficacy of TKS, particularly emphasizing its capabilities in handling temporal knowledge. TKS also endeavors to address temporal consistency issues, allowing the model to support a wide disparity between training and testing time steps. Consequently, after ensuring performance during the training phase with more extensive time steps, our algorithm can be deployed using fewer time steps. This adaptability fosters a conducive environment for deploying SNNs on edge devices. It simplifies and accelerates deployment on such devices, eliminating the need to retrain models when switching devices, and substantially curtails the latency typically associated with spiking neural networks.

In future studies, exploring how to leverage TKS for deploying SNNs on edge devices promises to be a fruitful research direction. TKS has the potential to aid hardware in reducing both latency and power consumption, which are critical requirements for edge devices. On the other hand, employing a strategy of temporal information transfer serves as a potent mechanism to enhance the performance of SNNs. Thus, another avenue for research is to delve deeper into the underlying mechanisms of this hidden knowledge and its distribution across different temporal points. This could lead to the development of more refined knowledge transfer strategies, optimizing the time steps between information flows.

\bibliographystyle{IEEEtran} 

\bibliography{elsarticle-template-harv}

\end{document}